\documentclass[10pt, a4paper]{article}

\usepackage[final]{lrec2026} 
\usepackage{linguex}
\usepackage{graphicx}
\usepackage{xcolor}
\usepackage{color}
\usepackage{tabularx}

\title{Evaluating the Impact of LLM-Assisted Annotation in a Perspectivized Setting: the Case of FrameNet Annotation}

\name{Frederico Belcavello\textsuperscript{1}, Ely Matos\textsuperscript{1}, Arthur Lorenzi\textsuperscript{1}, Lisandra Bonoto\textsuperscript{1}, Lívia Ruiz\textsuperscript{1},\\
{\bf \large Luiz Fernando Pereira\textsuperscript{1}, Victor Herbst\textsuperscript{1}, Yulla Navarro\textsuperscript{1}, Helen de Andrade Abreu\textsuperscript{1},}\\
{\bf \large Lívia Dutra\textsuperscript{1,2}, Tiago Timponi Torrent\textsuperscript{1,3}}}

\address{\textsuperscript{1} Federal University of Juiz de Fora | FrameNet Brasil,\\
\textsuperscript{2} Gothenburg University,\\
\textsuperscript{3} Brazilian National Council for Scientific and Technological Development - CNPq \\
         \{fred.belcavello, ely.matos, tiago.torrent\}@ufjf.br, \{arthur.lorenzi, lisandra.bonoto,\\
          livia.padua, luizfernando.pereira, victor.herbst, yulla.liquer\}@estudante.ufjf.br,\\
         livia.vicente.dutra@svenska.gu.se, helen.abreu@visitante.ufjf.br\\}

\abstract{
The use of LLM-based applications as a means to accelerate and/or substitute human labor in the creation of language resources and dataset is a reality. Nonetheless, despite the potential of such tools for linguistic research, comprehensive evaluation of their performance and impact on the creation of annotated datasets, especially under a perspectivized approach to NLP, is still missing. This paper contributes to reduction of this gap by reporting on an extensive evaluation of the (semi-)automatization of FrameNet-like semantic annotation by the use of an LLM-based semantic role labeler. The methodology employed compares annotation time, coverage and diversity in three experimental settings: manual, automatic and semi-automatic annotation. Results show that the hybrid, semi-automatic annotation setting leads to increased frame diversity and similar annotation coverage, when compared to the human-only setting, while the automatic setting performs considerably worse in all metrics, except for annotation time.
 \\ \newline \Keywords{FrameNet, LLM-assisted annotation, evaluation} 
}

\begin{document}

\maketitleabstract

\section{Introduction}

FrameNet is the implementation of the theory of Frame Semantics \citep{Fillmore1982} in the form of a language resource that clusters together lexical units whose meaning require the same background scene, called a frame, to be evoked for their comprehension \citep{Fillmore2003}. Annotation serves as the empirical backbone of FrameNet, providing the evidence necessary to support the conceptual and linguistic analysis within the FrameNet model: it both validates how Lexical Units (LUs),  defined as the pairing of a word with a specific frame, instantiate the frames they evoke, and yields semantically labeled datasets, which can be further used in several applications.\footnote{This paper is currently under peer review for a conference.}

FrameNet-style semantic annotation remains an essential yet labor intensive process. Since its creation \citep{baker-etal-1998-berkeley-framenet}, FrameNet has relied on meticulous manual curation, requiring trained linguists to identify frame-evoking elements and label their corresponding frame elements in context. This fine-grained approach has produced a high-quality resource, but at the cost of scalability: expanding coverage across domains and languages demands considerable amounts of human effort and time. Early studies on FrameNet-based semantic role labeling \citep{gildea-jurafsky-2000-automatic} have already pointed out that the lack of large annotated corpora constrained model performance.

Recent advances in large language models (LLMs) have opened new possibilities to reduce human workload in annotation tasks. Various research studies claim that conversational models demonstrate strong zero- and few-shot performance in annotation tasks \citep{brown2020language}. Others compare the performance of the model to that of crowd workers in different annotation tasks \citep{gilardi2023chatgpt}.

\begin{figure}
    \centering
    \includegraphics[width=1\linewidth]{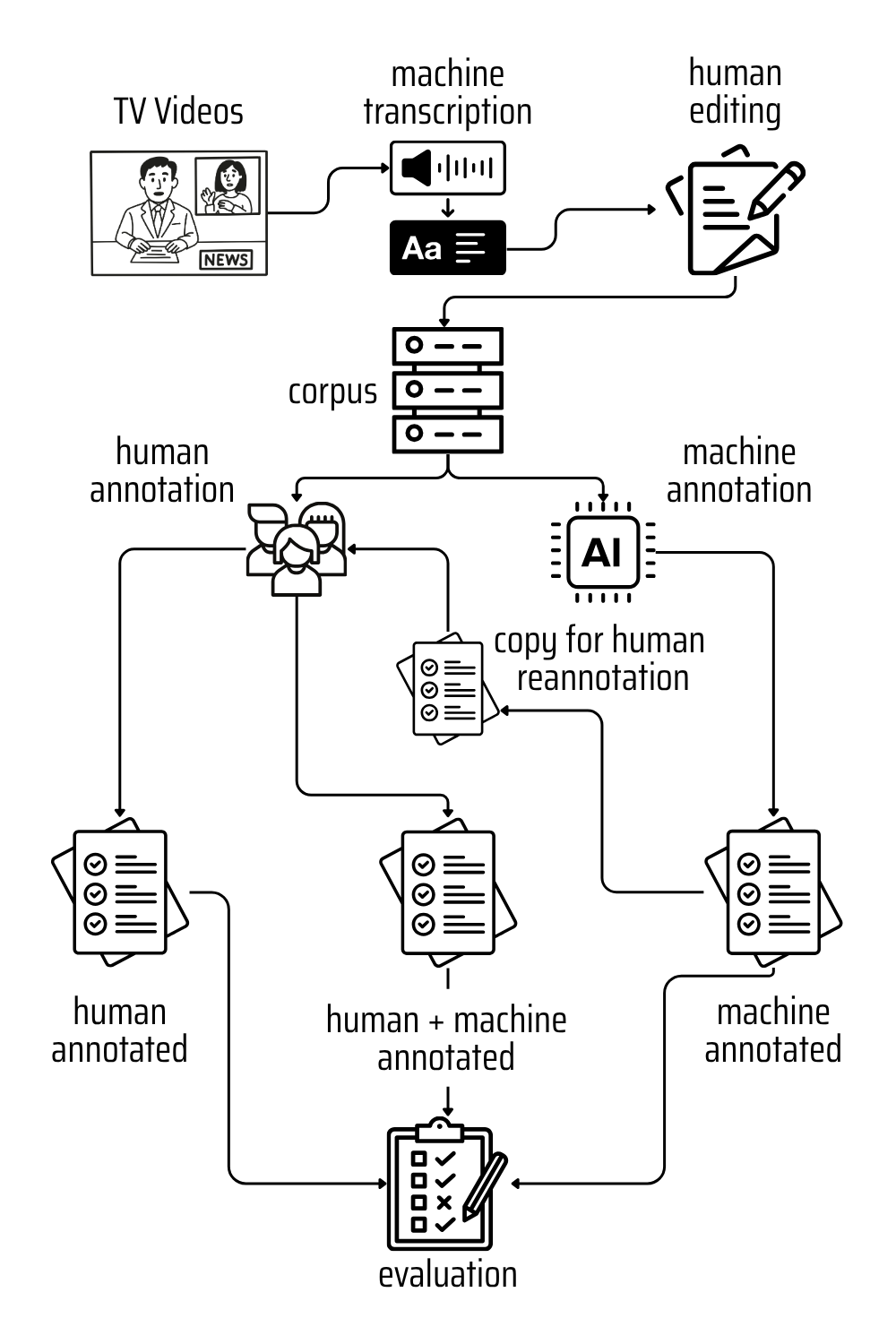}
    \caption{Experiment design}
    \label{fig:experimentdesign}
\end{figure}

Despite their potential, LLM-based annotation systems also pose significant risks. \citet{baumann2025largelanguagemodelhacking} show that subtle prompt or configuration changes can distort labels and introduce biases, a phenomenon they call LLM hacking. In large-scale tests, even state-of-the-art models produced incorrect or misleading annotations in roughly one third of cases. Such errors can propagate into downstream analyses, leading to false or exaggerated findings. The results by \citet{baumann2025largelanguagemodelhacking} highlight the need for rigorous human oversight and validation in LLM-assisted settings.

The question under investigation in this paper is whether LLMs can be used as assistants for facilitating, accelerating, and improving the quality of FrameNet annotation. Integrating these capabilities into FrameNet workflows could significantly lower the annotation barrier, aligning FrameNet with contemporary data-driven NLP while preserving its linguistic depth and interpretability. This semi-automatic approach, leveraging LLM-based automated annotation with human validation, may represent a viable path toward sustainable, large-scale FrameNet growth in the era of generative AI.

In this paper, we present an experiment designed to explore the semi-automation of FrameNet annotation through the use of LOME (Linguistically-Oriented Meaning Representation), an LLM-based open-source frame-semantic parser \citep{xia2021lome}. Our approach integrates LOME-generated suggestions into the human annotation workflow, allowing annotators to validate, correct, refine or delete automatically proposed frame and frame element labels, as shown in Figure \ref{fig:experimentdesign}. We hypothesize that such semi-automatic annotation can accelerate the FrameNet annotation process while maintaining quality. Beyond time efficiency, we also investigate whether machine-assisted annotation influences the diversity of frame interpretations and the perspectivization inherent to FrameNet annotation. Because FrameNet frames often encode viewpoint and conceptual stance, evaluating how automatic suggestions interact with these perspectival aspects is essential for understanding the epistemological impact of AI in the construction of human-curated language resources and datasets.

\section{FrameNet Annotation: a Perspectivized Approach to Semantic Role Labeling}
\label{sec:anno}

Frame Semantics \citep{Fillmore1982} proposes that linguistic meaning emerges from our ability to interpret expressions against structured conceptual backgrounds known as frames. A frame represents a schematic scene, such as a commercial transaction, a motion event, or a perception event, along with its participants, props, and internal relations. Words evoke these frames, activating the knowledge structures necessary for comprehension.

Within this theoretical model, perspective plays a central role. Language does not simply encode objective situations; it construes them from specific points of view \citep{trott-etal-2020-construing}. Fillmore's well-known contrast between land and ground, for instance, illustrates how two lexemes referring to the same physical region differ in conceptual vantage point: land is construed taking the sea as a reference point, while ground is construed having air as the opposing concept. In FrameNet, such perspectival distinctions are formally represented through frame-to-frame relations, for example, \texttt{Commerce\_buy} and \texttt{Commerce\_sell} are distinct perspectives on the more general \texttt{Commerce\_goods-transfer} frame.

The Berkeley FrameNet project \citep{Fillmore2003} implemented these ideas computationally by building a lexicon organized around frames and their frame elements (FEs), which correspond to the participants and attributes of each situation. For each frame, the database records: (i) the set of core and non-core FEs in each frame; (ii) the lexical units (LUs), pairings of words and senses that can evoke the frame; and (iii) a network of frame-to-frame relations that encode conceptual dependencies such as inheritance or perspective. This whole data structure is based on corpus evidence obtained from annotation.

There are two types of text annotation in FrameNet: lexicographic and full-text. In lexicographic annotation, sentences are queried in corpora with the aim of attesting the syntactic and semantic affordances of a given LU, which is known to evoke a given frame. It is common that, in the process of selecting examples to be annotated, annotators choose those where the frame being annotated for is clearly represented. On the other hand, in full-text annotation, the goal is to annotate every LU in a given text. Annotators should read the text and select, for every lexeme which is a potential annotation target, the appropriate frame.  Each LU thus yields its own Annotation Set (AS), so that a single sentence often generates multiple ASs targeting different LUs. This method places the focus on the broader discourse context, ensuring that all frame-evoking expressions in the text are represented. Because Frame Semantics assumes that different lexical items referring to the same conceptual scene impose distinct perspectives, full-text annotation naturally reveals these perspectival contrasts within coherent discourse. Lexicographic annotation, by contrast, is restricted to sentences selected to exemplify the valence patterns of a single predetermined LU, which limits the emergence of cross-frame perspective relations within the same text.

Crucially, FrameNet annotation differs from other approaches to semantic role labeling by treating meaning as interpretive rather than categorical. Because frames are prototype-based and overlap, a single expression may evoke more than one plausible frame depending on context. For example, in \ref{ex:children}, the noun \textit{school} may instantiate either the Goal FE in the \texttt{Motion} frame or the Activity FE in the \texttt{Activity\_start} frame, depending on how one interprets the semantics of \textit{going} in the sentence. Instead of enforcing a single correct label, the annotation process acknowledges that such variation reflects legitimate differences in interpretation, making FrameNet annotation inherently perspectivized \citep{cabitza2023toward}. 

\ex.\label{ex:children} Children who started going to school this year may have jobs that are yet to be invented.

From this standpoint, FrameNet annotation is not merely a task of tagging arguments, but an act of making explicit the interpretive stance that language adopts toward experience. This perspectivized understanding of semantic roles underlies the epistemological strength of the FrameNet model: it reveals how meaning is structured, rather than merely what is labeled. In the context of semi-automatic annotation, preserving this interpretive dimension is essential. By emphasizing perspective as an integral part of meaning representation, FrameNet offers both a theoretical and methodological foundation for the semi-automatic annotation approach explored in this study, one that combines the scalability of computational methods with the interpretive rigor of expert linguistic analysis.

\section{Related Work}
\label{sec:related}

Recent research has explored the use of LLM-based tools for a series of FrameNet-related tasks.

\citet{Torrent_Hoffmann_Almeida_Turner_2024} propose and evaluate a series of prompts for using conversational LLMs for the creation of LUs and frames. The authors recognize the potential of LLMs to augment FrameNet coverage and expand it to other languages. However, they do not explore frame annotation.

\citet{cui-swayamdipta-2024-annotating}, in turn, propose a methodology for generating synthetic annotated sentences from original human annotated ones. They claim that their method can be of use in low-resource settings but recognize its limitations in substituting human-annotated data where available.

Another group of methods approaches FrameNet annotation more directly. \citet{chundru2025llmsencodeframesemantics},  \citet{devasier2025llmsextractframesemanticarguments} and \citet{garat2025exploringincontextlearningframesemantic} propose methods to leverage conversational LLMs for that purpose. However, all three proposals require the frame and the frame elements to be included in the prompt along with the exemplar annotations. This type of methodology does not solve the problem of full-text annotation and requires sentences to be annotated to be previously sorted out and organized by frame-evoking targets.

Finally, several frame-semantic role labelers have been proposed over the years, using different computational techniques, as well as requiring different levels of complexity for their training and deployment, over the years \citep{das-etal-2014-frame,hartmann-etal-2017-domain,swayamdipta2017framesemanticparsingsoftmaxmarginsegmental,kalyanpur2020opendomainframesemanticparsing,jiang-riloff-2021-exploiting,xia2021lome,tamburini-2022-combining,an-etal-2023-coarse}. Despite assessing the performance of each system, these papers did not evaluate the incorporation of their frame parsers into an annotation pipeline. Therefore, they are not able to provide any evaluation of the impact of LLM assistance to the annotator's work.

To the best of our knowledge, the only paper that studied the extent to which semi-automatization impacts annotation is the one by \citet{rehbein-etal-2009-assessing}. In their paper, authors evaluate the impact of automatically pre-annotating sentences for frames and frame elements in terms of the amount of time needed to perform the annotations and the precision, recall and f-score for the annotations. \citet{rehbein2009assessing} evaluated three conditions of semi-automatization: one in which no pre-annotation was performed and annotators performed the whole process, one in which a state-of-the-art frame and FE labeler was used to pre-annotate the sentences, and a third in which errors were manually inserted in a gold-standard annotation. All three conditions were tested on lexicographic annotation.

\citet{rehbein2009assessing} concluded that pre-annotation did not have a statistically significant effect on annotation time, despite the fact that annotators were faster under the third annotation condition than when presented with the unannotated sentences. As for annotation quality, authors conclude that pre-annotation does improve the overall quality of the annotations. Moreover, they investigate the influence of pre-annotation on the types of errors human annotators make, and conclude that the pre-annotation does not seem to corrupt human judgments, since annotators make the same kinds of deviate judgments in all three conditions, when compared to the gold standard annotation.

The evaluation presented in this paper differs from that by \citet{rehbein2009assessing} in several ways: (a) because we recoginze the perspectivized nature of FrameNet annotation, we do not compare the outcomes of either human-only, machine-only and hybrid annotation to one gold standard reference dataset; (b) instead, we focus on measuring the impacts of LLM-assisted pre-annotation on the coverage, diversity, observance of minimal requirements and time spent on human annotation; and (c) we conduct the evaluation on a full-text annotation task, which yields the kind of annotated dataset that is more useful for downstream tasks. In the next section, we detail the experiment design used.

\section{Experiment Design}
\label{sec:design}

The experiment whose results are reported in this paper aimed to evaluate the potential of LLM-based tools for semi-automating FrameNet annotation. 

The experiment was structured in two phases. In Phase 1, each annotator manually annotated their assigned sentences from scratch, following the aforementioned guidelines. In Phase 2, each annotator was provided with a new set of sentences, different from those annotated in Phase 1, that had been automatically annotated by LOME and asked to revise it. Figure \ref{fig:experimentdesign} summarizes the experiment design, which is detailed in the following paragraphs.

\paragraph{Corpus} The experiment used sentences from a corpus of television news media in Brazilian Portuguese. The corpus comprises a total of 178 documents and 3,442 sentences. A subset of the corpus containing 12 documents and a total of 311 sentences was randomly selected. Three versions of this subset were created: (i) one for Human annotation, (ii) one for Machine annotation, and (iii) a copy of the product of the machine annotated data for the  Machine plus Human annotation.

\paragraph{Human Annotation} A group of five annotators, each with intermediate experience in FrameNet annotation, participated in the study. Each annotator was assigned approximately 60 sentences, distributed over two or three different news stories from the corpus. All annotations included frame and FE assignment, following FrameNet's guidelines for full-text annotation \citep{ruppenhofer2016framenet} and the FrameNet Brasil guidelines for multimodal annotation of audio-oriented videos \citep{belcavello2023}. The former indicates that annotators should watch each news story while annotating the sentences as a way of making sure that they pick the frame considering the multimodal context. The latter means that annotator creates ASs for each lexeme for which there is a LU in FrameNet, sentence by sentence.

\paragraph{Machine Annotation}

Machine annotation was done using LOME \citep{xia2021lome}. The choice for LOME is justified because it combines three important features: first, it is built upon an LLM, meaning that it leverages the potential of such models for frame semantic role labeling; second, it does not require any preprocessing of the sentences do be annotated to be made before they are submitted to the system, as it is the case with the conversational LLMs reviewed in section \ref{sec:related}; and, third, it can be trained for any language for which there are corpora annotated for FrameNet categories. 

For the experiment, we used a version of LOME trained on existing full-text annotation in both Brazilian Portuguese and English. In total, 18,170 annotated sentences were used––12,240 in Brazilain Portuguese and 5,930 in English––, with a rough average of 5.5 ASs per sentence \citep{dutra2024evaluating}.

The pipeline for the annotation of each sentence comprises the following steps:

\begin{enumerate}
    \item The sentence is parsed using the Trankit UD parser \citep{nguyen2021trankit}. This allows for the extraction of the tokens to be used as input to the next step. Each token has an associated lemma and a part of speech (POS).
    \item The tokens are inputted into LOME for processing. The result is a variable number of sets of frames and respective FEs, when assigned. For each frame or FE we have also the text span from the sentence where either the target LU or the linguistic material instantiating the FE is located.
    \item Using the position of text span, the system recovers the lemma/POS associated to each word in the span.
    \item Using lemma/POS and the attributed frame, the system checks if there is a LU already created. If not, a new LU is added, evoking the frame.
    \item Finally, with the LU, the system  creates a new AS, indicating the FEs automatically assigned using the span defined by LOME.
\end{enumerate}

The ASs created are associated with two copies of each sentence: one that joins the Machine annotation subset of the corpus, and another that serves as starting point for the Machine plus Human annotation. The latter will be edited by the human annotators and the former is preserved for comparison.

\paragraph{Machine plus Human Annotation} Each annotator received a new lot of sentences, different from the ones they annotated in the first phase. This time, the sentences were pre-annotated by LOME. Annotators were instructed to review and correct these LOME-generated labels, focusing on the adequacy of frame and frame element identification. 

The annotator had the following options (with their corresponding statuses): (a) fully accept the automatic annotation without making any corrections (ACCEPTED); (b) completely reject the machine annotation by removing the AS (DELETED); (c) replace the frame suggested in the Machine annotation while keeping the same lemma, or accept the LU suggested by the Machine annotation but modify (add or remove) some or all of the FEs (UPDATED); and finally, (d) create new ASs (CREATED).

This design allowed us to compare human and machine plus human annotation not only in terms of speed but also in terms of adherence to FrameNet methodology and diversity. We present the evaluation metrics used in the following section.

\section{Evaluation Metrics}

The methodology created to compare the three annotation configurations––human only, machine only and machine plus human––in the experiment sought to address two relevant aspects: the product and the process of annotation.

The evaluation of the annotation \textbf{product} consisted of counting how many elements were annotated (manually and automatically) and comparing these annotations. These elements include documents, sentences, annotation sets (ASs), and frame elements (FE). Still addressing the product aspect of the annotation and aiming to identify possible impacts of each annotation configuration on the diversity of labels used, we measured how many unique frames are associated with each document for each type of annotation, as well as the average number of frames occurring in each annotated sentence under each configuration. Finally, to qualitatively compare the annotation product, the cosine similarity between the semantic representations of each sentence \citep{viridiano-etal-2024-framed} was used. Once again, the comparison was carried out pairwise for all possible combinations of the three annotation configurations.

Regarding the annotation \textbf{process}, three measures were evaluated: (i) minimal number of core FEs in the annotation, (ii) the time spent by the annotators in the annotation, and (iii) the types of edits made to correct the automatic annotation in the machine plus human condition. 

Core FEs are those that directly express the semantics of the frame and they should all be present in the annotation, except in two cases: (a) when one core FE is in a \textbf{excludes} relation with one or more FEs, and (b) when two or more FEs are in a core set. The \texttt{Self\_motion} frame in \ref{ex:selfmotion} exemplifies both cases. 

\ex.\label{ex:selfmotion} \texttt{Self\_motion} \\
\textbf{Definition}: The \colorbox{red}{\textcolor{white}{Self\_mover}}, a living being, moves under its own direction along a \colorbox{blue}{\textcolor{white}{Path}}. Alternatively or in addition to \colorbox{blue}{\textcolor{white}{Path}}, an \colorbox{brown}{\textcolor{white}{Area}}, \colorbox{gray}{\textcolor{white}{Direction}}, \colorbox{teal}{\textcolor{white}{Source}}, or \colorbox{cyan}{\textcolor{white}{Goal}} for the movement may be mentioned.\\
\textbf{Core Frame Elements}:\\
\colorbox{brown}{\textcolor{white}{Area}}: It is used for expressions which describe a general area in which motion takes place when the motion is understood to be irregular and not to consist of a single linear path.\\
\textit{Excludes}: Direction, Goal, Path, Source.\\ 
\colorbox{gray}{\textcolor{white}{Direction}}: The direction that the \colorbox{red}{\textcolor{white}{Self\_mover}} heads in during the motion.\\ 
\colorbox{cyan}{\textcolor{white}{Goal}}: It is used for any expression which tells where the \colorbox{red}{\textcolor{white}{Self\_mover}} ends up as a result of the motion.\\
\colorbox{blue}{\textcolor{white}{Path}}: It is used for any description of a trajectory of motion which is neither a \colorbox{teal}{\textcolor{white}{Source}} nor a \colorbox{cyan}{\textcolor{white}{Goal}}.\\
\colorbox{red}{\textcolor{white}{Self\_mover}}: It is the living being which moves under its own power.\\
\colorbox{teal}{\textcolor{white}{Source}}: It is used for any expression which implies a definite starting-point of motion.\\
\textbf{FE Core Set(s)}: \{Source, Goal, Path, Direction\}\\

Note that, although the \texttt{Self\_motion} frame has eight core FEs, there can be one complete annotation for this frame with only two FEs: Self\_mover and Area or Self\_mover and either Source, Path, Goal or Direction. This is so, first, because the presence of the Area FE excludes the possibility of any of the other locative FEs, as the odd example in \ref{ex:areaannot} demonstrates. Second, because Source, Path, Goal and Direction are in a core set, only one of them must be present for the frame to instantiate, as \ref{ex:spgannot} shows: the sentence is well formed with only one of the FEs following the verb, with any combination of two of them and also with all three of them.   

\ex.\label{ex:areaannot}\colorbox{red}{\textcolor{white}{Mark}} was  \textbf{running} \colorbox{brown}{\textcolor{white}{around}} \colorbox{cyan}{\textcolor{white}{to school}}.

\ex.\label{ex:spgannot}\colorbox{red}{\textcolor{white}{Mark}} was  \textbf{running} \colorbox{teal}{\textcolor{white}{from home}} \colorbox{cyan}{\textcolor{white}{to school}} \colorbox{blue}{\textcolor{white}{along the road}}.

Hence, the minimum number of FEs that must necessarily be annotated was calculated by taking into consideration the total number of FEs in a frame minus the ones in exclude relations and core sets, where only one of the possible FEs was counted. The percentage of core FEs annotated indicates how complete an annotation is. 

The time measure was taken for each AS. In the case of human annotation, this measure considered the time spent by the annotator both to define a new AS to be annotated––by selecting the appropriate LU––and to record the FEs in the AS. For the editing of the LOME output under the Machine plus Human configuration, the ASs had already been created, thus the recorded time refers to the edits (or removals) made to each AS. 

Another measure related to the annotation process concerns the edits made to the Machine Annotation during the Machine plus Human annotation, according to the possibilities described in section \ref{sec:design}. 

\section{Results and discussion}
\label{sec:results}

As indicated in section \ref{sec:design}, a total of 12 documents comprising 311 sentences were used in the experiment. Each sentence has a variable number of associated ASs. An AS indicates the Lexical Unit (LU) associated with an expression in the sentence and, at the same time, the frame evoked by that LU. In turn, each AS is associated with a variable number of FEs. On average, the number of AS per document was 129 in the Human annotation condition, 126 in the Machine annotation, and 160 in the Machine plus Human annotation. This first measure reveals that, while the number of ASs varies very little in the comparison between the human-only and the machine-only conditions, it increases sensibly in the human plus machine condition, presenting a 24\% increase when compared to the human-only scenario and a 26.9\% increase when compared to the machine-only setting. 

\begin{table*}[!ht]
\begin{center}
\small
\begin{tabular}{|l|c|c|c|c|c|c|c|c|c|c|c|c|}
\hline
 & \multicolumn{3}{c|}{Human Annotation} & \multicolumn{3}{c|}{Machine Annotation} & \multicolumn{3}{c|}{Machine + Human Annotation} \\
\hline
Doc & Sent & Frames & Avg F/S & Sent & Frames & Avg F/S & Sent & Frames & Avg F/S \\
\hline
02\_13 & 22 & 71 & 3.23 & 22 & 53 & 2.41 & 22 & 80 & 3.64 \\
02\_14 & 13 & 71 & 5.46 & 23 & 56 & 2.43 & 23 & 105 & 4.57 \\
03\_11 & 27 & 77 & 2.85 & 26 & 56 & 2.15 & 28 & 88 & 3.14 \\
03\_12 & 19 & 47 & 2.47 & 26 & 57 & 2.19 & 27 & 85 & 3.15 \\
04\_01 & 50 & 114 & 2.28 & 46 & 87 & 1.89 & 49 & 99 & 2.02 \\
04\_06 & 9 & 34 & 3.78 & 9 & 20 & 2.22 & 10 & 26 & 2.60 \\
05\_01 & 14 & 54 & 3.86 & 15 & 26 & 1.73 & 15 & 51 & 3.40 \\
05\_02 & 26 & 80 & 3.08 & 23 & 54 & 2.35 & 26 & 93 & 3.58 \\
05\_03 & 3 & 12 & 4.00 & 20 & 59 & 2.95 & 20 & 83 & 4.15 \\
07\_02 & 21 & 97 & 4.62 & 22 & 58 & 2.64 & 22 & 104 & 4.73 \\
07\_03 & 13 & 80 & 6.15 & 13 & 46 & 3.54 & 13 & 75 & 5.77 \\
07\_07 & 20 & 78 & 3.90 & 17 & 60 & 3.53 & 20 & 82 & 4.10 \\
\hline
Avg & 19.75 & 67.91 & 3.80 & 21.83 & 52.66 & 2.50 & 22.91 & 80.91 & 3.74 \\
\hline
\end{tabular}
\caption{Frame diversity across documents}
\label{tab:diversidade-frames}
\end{center}
\end{table*}

Beyond the average number of ASs per sentence, we also measured the frame diversity in each condition. The motivation behind this metric is to assess whether the use of LLMs could interfere with human judgment when annotating, similarly to what is investigated by \citet{rehbein2009assessing}. Table \ref{tab:diversidade-frames} shows the number of unique frames used in each annotation setting.\footnote{The number of sentences in each document varies across annotation conditions because, in the comparison, only sentences with at least one annotation set are considered. Variations in the number of sentences can indicate that either LOME was not able to find any frames for a given sentence, or that a human annotator did not associate any frame to a sentence, for any reason.} A higher number of unique frames may be indicative that not only more AS were created––which was the case for the Human plus Machine condition––but also that different perspectives were adopted by annotators for one same lexeme, and, therefore, that more possibilities provided by the vast number of annotation labels available in FrameNet were used.\footnote{The FrameNet Brasil database \citep{torrent-et-al-2022}, which was used in this experiment, offers 1,429 different frames and 13,071 different FEs for annotation.}   Naturally, those two aspects are interconnected. Data presented in Table \ref{tab:diversidade-frames} indicate that the Machine annotation is still very limited in terms of frame diversity. Considering that, on average, the number of frames per document in the Human annotation and in the Machine annotation is very similar, the difference in the average number of \textbf{unique} frames per document––67.91 for the Human annotation versus 52.66 for the Machine annotation––and per sentence––3.80 for the Human annotation versus 2.50 for the Machine annotation––is considerable. Moreover, the Machine plus Human condition led to a higher average of unique frames per document and a similar one per sentence when compared with the Human annotation scenario. 

\begin{table*}[!ht]
\begin{center}
\small
\begin{tabular}{|l|c|c|c|}
\hline
Doc & Human vs Machine & Human vs Machine + Human & Machine vs Machine + Human \\
\hline
02\_13 & 0.7199	 & 0.7763	 & 0.8461 \\
02\_14 & 	0.6918	 & 0.8267	 & 0.8509 \\
03\_11 & 	0.5625	 & 0.7768	 & 0.6927 \\
03\_12 & 	0.6153	 & 0.7288	 & 0.7547 \\
04\_01 & 	0.5686	 & 0.6757	 & 0.8322 \\
04\_06 & 	0.5982	 & 0.7080	 & 0.7668 \\
05\_01 & 	0.6167	 & 0.7193	 & 0.7520 \\
05\_02 & 	0.6672	 & 0.7264	 & 0.8175 \\
05\_03 & 	0.6835	 & 0.7250	 & 0.9116 \\
07\_02 & 	0.6182	 & 0.7752	 & 0.7106 \\
07\_03 & 	0.6053	 & 0.7843	 & 0.7186 \\
07\_07 & 	0.6370	 & 0.7355	 & 0.7279 \\
\hline
Avg  & 	0.6320	 & 0.7465	 & 0.7818 \\
\hline
\end{tabular}
\caption{Cosine similarity between annotation methods}
\label{tab:cosine-similarity}
\end{center}
\end{table*}

Table \ref{tab:cosine-similarity} can be broadly understood as representing the degree of agreement regarding the semantic representation of each sentence in each document between the three types of annotation. Although the automatic annotation identified fewer frames, the data suggest that these frames were mostly kept in the Machine plus Human annotation condition. On the other hand, the also high similarity between the Human and the Machine plus Human conditions indicates that human judgements were preserved in the pre-annotated configuration. This data will be reinforced by the findings to the discussed in the end of this section.  

\begin{table*}[!ht]
\begin{center}
\small
\begin{tabular}{|l|c|c|c|c|c|c|c|c|c|c|c|c|}
\hline
 & \multicolumn{3}{c|}{Human Annotation} & \multicolumn{3}{c|}{Machine Annotation} & \multicolumn{3}{c|}{Machine + Human Annotation} \\
\hline
Doc & Core & Min & \% & Core & Min & \% & Core & Min & \% \\
\hline
02\_13 & 229 & 246 & 93.09 & 92 & 244 & 37.70 & 338 & 299 & 100.00 \\
02\_14 & 309 & 301 & 100.00 & 120 & 352 & 34.09 & 523 & 491 & 100.00 \\
03\_11 & 268 & 296 & 90.54 & 106 & 267 & 39.70 & 350 & 384 & 91.15 \\
03\_12 & 193 & 248 & 77.82 &  111 & 360 & 30.83 & 358 & 412 & 86.89 \\
04\_01 & 578 & 624 & 92.63 & 186 & 543 & 34.25 & 407 & 567 & 71.78 \\
04\_06 & 159 & 128 & 100.00 & 27 & 99 & 27.27 & 130 & 152 & 85.53 \\
05\_01 & 210 & 194 & 100.00 & 49 & 132 & 37.12 & 173 & 213 & 81.22 \\
05\_02 & 447 & 332 & 100.00 & 90 & 296 & 30.41 & 346 & 428 & 80.84 \\
05\_03 & 38 & 29 & 100.00 & 107 & 291 & 36.77 & 284 & 314 & 90.45 \\
07\_02 & 438 & 400 & 100.00 & 108 & 321 & 33.64 & 465 & 386 & 100.00 \\
07\_03 & 294 & 308 & 95.45 & 66 & 197 & 33.50 & 318 & 284 & 100.00 \\
07\_07 & 323 & 312 & 100.00 & 87 & 248 & 35.08 & 364 & 314 & 100.00 \\
\hline
Avg & 290.5 & 284.83 & 95.79 & 95.75 & 279.17 & 34.20 & 338 & 353.67 & 90.65 \\
\hline
\end{tabular}
\caption{Percentage of core FEs annotated}
\label{tab:fes-core-anotados}
\end{center}
\end{table*}

Table \ref{tab:fes-core-anotados} shows the percentage of the minimal number of core FEs annotated per document. This metric assesses whether the annotation respected FrameNet methodology regarding the requirement that core FEs are present or inferred in the sentence for a frame to be instantiated. Data shows that human annotators excel in following the guidelines, with 95.79\% of the minimal number of FEs being annotated. On the contrary, LOME falls short in this metric, with only 34.20\% of the minimal number of FEs present in the annotation. This is mainly due to the fact that LOME does not indicate inferrable FEs––what FrameNet calls null instantiations \citep{ruppenhofer2016framenet}––in the annotation. This is so because, as noted in section \ref{sec:design}, LOME needs to assign FEs to spans of text in the sentence. In the Machine plus Human configuration, the percentage of minimal core FEs annotated decreases to 90.65\%, which may be due some minor influence of the pre-annotation on the human annotator's judgment about the adequacy of the annotation to the FrameNet policy. This effect does not seem to be very relevant, though.

\begin{table*}[!ht]
\begin{center}
\small
\begin{tabular}{|l|c|c|c|c|c|}
\hline
Doc & Sent & Avg Length & Human Anno & Machine +Human Anno & Diff \\
\hline
02\_13 & 20 & 82.1 & 9.36 & 9.37 & -0.1 \\
02\_14 & 14 & 152.64 & 19.01 & 11.03 & 7.98 \\
03\_11 & 26 & 101.58 & 4.57 & 9.89 & -5.32 \\
03\_12 & 21 & 80.14 & 2.76 & 4.61 & -1.85 \\
04\_01 & 43 & 88.07 & 19.17 & 3.23 & 15.94 \\
04\_06 & 7 & 101.29 & 16.15 & 6.34 & 9.81 \\
05\_01 & 13 & 91.08 & 26.91 & 38.12 & -11.21 \\
05\_02 & 26 & 104.5 & 23.6 & 18.13 & 5.47 \\
05\_03 & 3 & 129.33 & 20.45 & 5.72 & 14.73 \\
07\_02 & 20 & 116.0 & 13.61 & 19.61 & -6 \\
07\_03 & 13 & 135.85 & 13.77 & 17.68 & -3.91 \\
07\_07 & 19 & 107.11 & 10.19 & 11.89 & -1.7 \\
\hline
Avg & 18.75 & 107.47 & 14.96 & 12.97 & 1.99 \\
\hline
\end{tabular}
\caption{Average annotation time per sentence in minutes}
\label{tab:tempo-medio-anotacao}
\end{center}
\end{table*}

The average time recorded for the annotation of each sentence in each document is presented in Table \ref{tab:tempo-medio-anotacao}. This metric only compares the Human-only condition with the Machine plus Human one. This is because the Machine-only annotation is performed very fast in comparison to the other two. For the total of the experiment it shows that using pre-annotation leads to a small decrease of average time per sentence: 1.99 minutes. This indicates that reducing annotation time seems to be not the most compelling argument in favor of using LLM-based pre-annotation. This conclusion replicates the results in \citet{rehbein2009assessing}.

\begin{table*}[!ht]
\begin{center}
\small
\begin{tabular}{|l|c|c|c|c|c|c|c|c|c|c|c|}
\hline
\footnotesize Doc & Total & ACCEPTED & \% & CREATED & \% & DELETED & \% & UPDATED & \% \\
\hline
03\_11 & 230 & 7 & 3.04 & 18 & 10.84 & 64 & 27.83 & 141 & 61.30  \\
03\_12 & 228 & 2 & 0.88 & 18 & 10.59 & 58 & 25.44 & 150 & 65.79  \\
02\_13 & 161 & 0 & 0.00 & 25 & 17.12 & 15 & 9.32 & 121 & 75.16  \\
02\_14 & 276 & 0 & 0.00 & 18 & 7.59 & 39 & 14.13 & 219 & 79.35  \\
04\_01 & 320 & 20 & 6.25 & 68 & 26.05 & 59 & 18.44 & 173 & 54.06  \\
04\_06 & 80 & 2 & 2.50 & 4 & 5.80 & 11 & 13.75 & 63 & 78.75  \\
05\_01 & 113 & 24 & 21.24 & 7 & 6.80 & 10 & 8.85 & 72 & 63.72  \\
05\_02 & 222 & 47 & 21.17 & 5 & 2.76 & 41 & 18.47 & 129 & 58.11  \\
05\_03 & 139 & 27 & 19.42 & 19 & 14.73 & 10 & 7.19 & 83 & 59.71  \\
07\_02 & 253 & 4 & 1.58 & 8 & 4.60 & 79 & 31.23 & 162 & 64.03  \\
07\_03 & 182 & 2 & 1.10 & 9 & 7.03 & 54 & 29.67 & 117 & 64.29  \\
07\_07 & 229 & 5 & 2.18 & 11 & 7.05 & 73 & 31.88 & 140 & 61.14 \\
\hline
Avg & 202.75 & 11.67 & 6.61 & 17.5 & 10.08 & 42.75 & 19.68 & 130.83 & 65.45\\
\hline
\end{tabular}
\caption{Human edits on LOME annotations}
\label{tab:edicoes-humanas-lome}
\end{center}
\end{table*}

Finally, Table \ref{tab:edicoes-humanas-lome} presents, in absolute and percentage terms, the edits annotators made to LOME pre-annotation during phase 2. Note that annotators completely discarded 19.68\% of automatic annotations and fully accepted only 6.61\% of them. This indicates that pre-annotation by LOME was far from being judged as perfect in this experimental setting. However, while 17.5\% of the annotation sets in the Machine plus Human condition were created from scratch, the majority of the ASs in the final dataset, 65.45\%, were partially used and improved by the annotators. This finding correlates to the one presented in Table \ref{tab:cosine-similarity}, since the similar cosine similarity between the Human and Machine plus Human conditions, and the one between the Machine and Machine plus Human conditions are indicative of partial preservation of both the frames obtained in the pre-annotation and the original human judgement of the sentences. 

Data from Table \ref{tab:edicoes-humanas-lome}, together with the ones presented in the previous tables, reinforces the idea that pre-annotation is valid as a strategy for increasing the number and diversity of ASs, while having little impact on both annotation time––with a small increase in performance––and on observance of FrameNet annotation guidelines––with a small decrease in the percentage of the minimal number of core FEs annotated.  

\section{Conclusions and outlook}

The experiment reported in this paper showed that LLM-based pre-annotation can be useful for improving the coverage of perspectivized FrameNet annotation, while preserving human judgment. Although no sensible improvement in annotation speed was observed, the use of pre-annotation validated by humans seems to be a viable path for fine-grained semantic annotation.

Future work should look at least into two extensions that should be adopted in the short term for a new evaluation of the impact of LLM assistants on FrameNet annotation. 

The first is the inclusion of other types of semantic role labelers––such as DAISY \citep{torrent2024flexible}, for example––in the automatic annotation pipeline. Additional parsers could serve as a post-processing step for LOME, aimed at adding annotations LOME was unable to perform.

A second aspect concerns the adoption of stricter policies in the annotation process to ensure that at least the minimum number of core FEs have been annotated (both in human and automatic annotation). This implies enabling the automatic system to record the occurrence of null instantiations when necessary.

\section{Acknowledgements}

Research reported in this paper was developed under the ReINVenTA––Research and Innovation Network for Vision and Text Analysis of Multimodal Objects––initiative, funded by the Minas Gerais State Agency for Research and Development (FAPEMIG – grant RED-00106-21) and the Brazilian National Council for Scientific and Technological Development (CNPq – grant 420945/2022-9). Belcavello was supported by CNPq (grant 200270/2023-0). Lorenzi, Abreu and Pereira were supported by FAPEMIG. Bonoto, Ruiz, Herbst and Navarro were supported by CNPq. Torrent is a research productivity grantee of CNPq (grant 311241/2025-5).  

\section{Ethical considerations and limitations}

All annotation used in the experiments, including for the annotation sets used for training the Brazilian Portuguese instance of LOME, was carried out by trained annotators who were paid a monthly stipend, which is, at least, equivalent to the minimum wage according to local regulations. All annotators involved in the annotation of the corpus used in the evaluation experiment reported here are co-authors of this paper.

Among the limitations of the experiment described, it is worth noting that all annotated sentences are written in Brazilian Portuguese. However, LOME is language-agnostic and its components are designed to prioritize multilinguality. LOME employs XLM-R \citep{conneau-etal-2020-unsupervised} as the underlying encoder, which allows the experiment to be easily extended to other languages. Furthermore, the experiment relied exclusively on LOME as the frame-semantic parser, but other LLM-based semantic role labelers may be evaluated in future work.

\section{Bibliographical References}
\label{sec:reference}

\bibliographystyle{lrec2026-natbib}
\bibliography{lrec2026-example,ACL_Anthology_part_aa,ACL_Anthology_part_ab,ACL_Anthology_part_ac}

\begin{thebibliography}{32}
\expandafter\ifx\csname natexlab\endcsname\relax\def\natexlab#1{#1}\fi

\bibitem[{An et~al.(2023)An, Zheng, Gao, Zhao, and Chang}]{an-etal-2023-coarse}
Kaikai An, Ce~Zheng, Bofei Gao, Haozhe Zhao, and Baobao Chang. 2023.
\newblock \href {https://doi.org/10.18653/v1/2023.findings-emnlp.897} {Coarse-to-fine dual encoders are better frame identification learners}.
\newblock In \emph{Findings of the Association for Computational Linguistics: EMNLP 2023}, pages 13455--13466, Singapore. Association for Computational Linguistics.

\bibitem[{Baker et~al.(1998)Baker, Fillmore, and Lowe}]{baker-etal-1998-berkeley-framenet}
Collin~F. Baker, Charles~J. Fillmore, and John~B. Lowe. 1998.
\newblock \href {https://doi.org/10.3115/980845.980860} {The {B}erkeley {F}rame{N}et project}.
\newblock In \emph{36th Annual Meeting of the Association for Computational Linguistics and 17th International Conference on Computational Linguistics, Volume 1}, pages 86--90, Montreal, Quebec, Canada. Association for Computational Linguistics.

\bibitem[{Baumann et~al.(2025)Baumann, Röttger, Urman, Wendsjö, del Arco, Gruber, and Hovy}]{baumann2025largelanguagemodelhacking}
Joachim Baumann, Paul Röttger, Aleksandra Urman, Albert Wendsjö, Flor Miriam~Plaza del Arco, Johannes~B. Gruber, and Dirk Hovy. 2025.
\newblock \href {http://arxiv.org/abs/2509.08825} {Large language model hacking: Quantifying the hidden risks of using llms for text annotation}.

\bibitem[{Belcavello(2023)}]{belcavello2023}
Frederico Belcavello. 2023.
\newblock \href {https://repositorio.ufjf.br/jspui/handle/ufjf/15527} {\emph{FrameNet annotation for multimodal corpora: Devising a methodology for the semantic representation of text-image interactions in audiovisual productions}}.
\newblock Doctoral dissertation, Universidade Federal de Juiz de Fora, Juiz de Fora, Brazil.
\newblock Faculdade de Letras, Programa de Pós-Graduação em Linguística. Advisor: Tiago Timpone Torrent. Co-advisor: Mark Turner.

\bibitem[{Brown et~al.(2020)Brown, Mann, Ryder, Subbiah, Kaplan, Dhariwal et~al.}]{brown2020language}
Tom~B. Brown, Benjamin Mann, Nick Ryder, Melanie Subbiah, Jared Kaplan, Prafulla Dhariwal, et~al. 2020.
\newblock Language models are few-shot learners.
\newblock In \emph{Advances in Neural Information Processing Systems (NeurIPS)}, volume~33, pages 1877--1901.

\bibitem[{Cabitza et~al.(2023)Cabitza, Campagner, and Basile}]{cabitza2023toward}
Federico Cabitza, Andrea Campagner, and Valerio Basile. 2023.
\newblock Toward a perspectivist turn in ground truthing for predictive computing.
\newblock In \emph{Proceedings of the AAAI Conference on Artificial Intelligence}, volume~37, pages 6860--6868.

\bibitem[{Chundru et~al.(2025)Chundru, Poddar, Cao, and Jiang}]{chundru2025llmsencodeframesemantics}
Jayanth~Krishna Chundru, Rudrashis Poddar, Jie Cao, and Tianyu Jiang. 2025.
\newblock \href {http://arxiv.org/abs/2509.19540} {Do llms encode frame semantics? evidence from frame identification}.
\newblock \emph{arXiv preprint arxiv: 2509.19540}.

\bibitem[{Conneau et~al.(2020)Conneau, Khandelwal, Goyal, Chaudhary, Wenzek, Guzm{\'a}n, Grave, Ott, Zettlemoyer, and Stoyanov}]{conneau-etal-2020-unsupervised}
Alexis Conneau, Kartikay Khandelwal, Naman Goyal, Vishrav Chaudhary, Guillaume Wenzek, Francisco Guzm{\'a}n, Edouard Grave, Myle Ott, Luke Zettlemoyer, and Veselin Stoyanov. 2020.
\newblock \href {https://doi.org/10.18653/v1/2020.acl-main.747} {Unsupervised cross-lingual representation learning at scale}.
\newblock In \emph{Proceedings of the 58th Annual Meeting of the Association for Computational Linguistics}, pages 8440--8451, Online. Association for Computational Linguistics.

\bibitem[{Cui and Swayamdipta(2024)}]{cui-swayamdipta-2024-annotating}
Xinyue Cui and Swabha Swayamdipta. 2024.
\newblock \href {https://doi.org/10.18653/v1/2024.acl-short.63} {Annotating {F}rame{N}et via structure-conditioned language generation}.
\newblock In \emph{Proceedings of the 62nd Annual Meeting of the Association for Computational Linguistics (Volume 2: Short Papers)}, pages 681--692, Bangkok, Thailand. Association for Computational Linguistics.

\bibitem[{Das et~al.(2014)Das, Chen, Martins, Schneider, and Smith}]{das-etal-2014-frame}
Dipanjan Das, Desai Chen, Andr{\'e} F.~T. Martins, Nathan Schneider, and Noah~A. Smith. 2014.
\newblock \href {https://doi.org/10.1162/COLI_a_00163} {Frame-semantic parsing}.
\newblock \emph{Computational Linguistics}, 40(1):9--56.

\bibitem[{Devasier et~al.(2025)Devasier, Mediratta, and Li}]{devasier2025llmsextractframesemanticarguments}
Jacob Devasier, Rishabh Mediratta, and Chengkai Li. 2025.
\newblock \href {http://arxiv.org/abs/2502.12516} {Can llms extract frame-semantic arguments?}
\newblock \emph{arXiv preprint arxiv: 2502.12516}.

\bibitem[{Dutra(2024)}]{dutra2024evaluating}
Lívia Dutra. 2024.
\newblock \href {https://gupea.ub.gu.se/handle/2077/81763} {Evaluating the contribution of framenet to gender-based violence identification: How semantic annotation can be used as a resource for identifying patterns of violence}.
\newblock Masters {T}hesis in {L}anguage {T}echnology, Göteborgs Universitet, Gothenburg.

\bibitem[{Fillmore(1982)}]{Fillmore1982}
Charles~J. Fillmore. 1982.
\newblock Frame semantics.
\newblock In {Linguistic Society of Korea}, editor, \emph{Linguistics in the Morning Calm}, pages 111--138. Hanshin Publishing Co., Seoul, South Korea.

\bibitem[{Fillmore et~al.(2003)Fillmore, Johnson, and Petruck}]{Fillmore2003}
Charles~J. Fillmore, Christopher~R. Johnson, and Miriam R.~L. Petruck. 2003.
\newblock Background to framenet.
\newblock \emph{International Journal of Lexicography}, 16(3):235--250.

\bibitem[{Garat et~al.(2025)Garat, Moncecchi, and Wonsever}]{garat2025exploringincontextlearningframesemantic}
Diego Garat, Guillermo Moncecchi, and Dina Wonsever. 2025.
\newblock \href {http://arxiv.org/abs/2507.23082} {Exploring in-context learning for frame-semantic parsing}.
\newblock \emph{arXiv preprint arxiv: 2507.23082}.

\bibitem[{Gilardi et~al.(2023)Gilardi, Alizadeh, and Kubli}]{gilardi2023chatgpt}
Fabrizio Gilardi, Meysam Alizadeh, and Maël Kubli. 2023.
\newblock \href {https://www.pnas.org/doi/abs/10.1073/pnas.2305016120} {Chat{GPT} outperforms crowd workers for text-annotation tasks}.
\newblock \emph{Proceedings of the National Academy of Sciences}, 120(30):e2305016120.

\bibitem[{Gildea and Jurafsky(2000)}]{gildea-jurafsky-2000-automatic}
Daniel Gildea and Daniel Jurafsky. 2000.
\newblock \href {https://doi.org/10.3115/1075218.1075283} {Automatic labeling of semantic roles}.
\newblock In \emph{Proceedings of the 38th Annual Meeting of the Association for Computational Linguistics}, pages 512--520, Hong Kong. Association for Computational Linguistics.

\bibitem[{Hartmann et~al.(2017)Hartmann, Kuznetsov, Martin, and Gurevych}]{hartmann-etal-2017-domain}
Silvana Hartmann, Ilia Kuznetsov, Teresa Martin, and Iryna Gurevych. 2017.
\newblock \href {https://aclanthology.org/E17-1045/} {Out-of-domain {F}rame{N}et semantic role labeling}.
\newblock In \emph{Proceedings of the 15th Conference of the {E}uropean Chapter of the Association for Computational Linguistics: Volume 1, Long Papers}, pages 471--482, Valencia, Spain. Association for Computational Linguistics.

\bibitem[{Jiang and Riloff(2021)}]{jiang-riloff-2021-exploiting}
Tianyu Jiang and Ellen Riloff. 2021.
\newblock \href {https://doi.org/10.18653/v1/2021.eacl-main.206} {Exploiting definitions for frame identification}.
\newblock In \emph{Proceedings of the 16th Conference of the European Chapter of the Association for Computational Linguistics: Main Volume}, pages 2429--2434, Online. Association for Computational Linguistics.

\bibitem[{Kalyanpur et~al.(2020)Kalyanpur, Biran, Breloff, Chu-Carroll, Diertani, Rambow, and Sammons}]{kalyanpur2020opendomainframesemanticparsing}
Aditya Kalyanpur, Or~Biran, Tom Breloff, Jennifer Chu-Carroll, Ariel Diertani, Owen Rambow, and Mark Sammons. 2020.
\newblock \href {http://arxiv.org/abs/2010.10998} {Open-domain frame semantic parsing using transformers}.
\newblock \emph{arXiv preprint arxiv: 2010.10998}.

\bibitem[{Nguyen et~al.(2021)Nguyen, Lai, Veyseh, and Nguyen}]{nguyen2021trankit}
Minh~Van Nguyen, Viet Lai, Amir Pouran~Ben Veyseh, and Thien~Huu Nguyen. 2021.
\newblock Trankit: A light-weight transformer-based toolkit for multilingual natural language processing.
\newblock In \emph{Proceedings of the 16th Conference of the European Chapter of the Association for Computational Linguistics: System Demonstrations}.

\bibitem[{Rehbein et~al.(2009{\natexlab{a}})Rehbein, Ruppenhofer, and Sporleder}]{rehbein-etal-2009-assessing}
Ines Rehbein, Josef Ruppenhofer, and Caroline Sporleder. 2009{\natexlab{a}}.
\newblock \href {https://aclanthology.org/W09-3003/} {Assessing the benefits of partial automatic pre-labeling for frame-semantic annotation}.
\newblock In \emph{Proceedings of the Third Linguistic Annotation Workshop ({LAW} {III})}, pages 19--26, Suntec, Singapore. Association for Computational Linguistics.

\bibitem[{Rehbein et~al.(2009{\natexlab{b}})Rehbein, Ruppenhofer, and Sporleder}]{rehbein2009assessing}
Ines Rehbein, Josef Ruppenhofer, and Caroline Sporleder. 2009{\natexlab{b}}.
\newblock Assessing the benefits of partial automatic pre-labeling for frame-semantic annotation.
\newblock In \emph{Proceedings of the Third Linguistic Annotation Workshop (LAW III)}, pages 19--27.

\bibitem[{Ruppenhofer et~al.(2016)Ruppenhofer, Ellsworth, Petruck, Johnson, and Scheffczyk}]{ruppenhofer2016framenet}
Josef Ruppenhofer, Michael Ellsworth, Miriam Petruck, Christopher Johnson, and Jan Scheffczyk. 2016.
\newblock \emph{Frame{N}et {II}: Extended Theory and Practice}.
\newblock International Computer Science Institute (ICSI).

\bibitem[{Swayamdipta et~al.(2017)Swayamdipta, Thomson, Dyer, and Smith}]{swayamdipta2017framesemanticparsingsoftmaxmarginsegmental}
Swabha Swayamdipta, Sam Thomson, Chris Dyer, and Noah~A. Smith. 2017.
\newblock \href {http://arxiv.org/abs/1706.09528} {Frame-semantic parsing with softmax-margin segmental rnns and a syntactic scaffold}.
\newblock \emph{arXiv preprint arxiv: 1706.09528}.

\bibitem[{Tamburini(2022)}]{tamburini-2022-combining}
Fabio Tamburini. 2022.
\newblock \href {https://aclanthology.org/2022.lrec-1.178/} {Combining {ELECTRA} and adaptive graph encoding for frame identification}.
\newblock In \emph{Proceedings of the Thirteenth Language Resources and Evaluation Conference}, pages 1671--1679, Marseille, France. European Language Resources Association.

\bibitem[{Torrent et~al.(2024{\natexlab{a}})Torrent, Hoffmann, Almeida, and Turner}]{Torrent_Hoffmann_Almeida_Turner_2024}
Tiago~Timponi Torrent, Thomas Hoffmann, Arthur~Lorenzi Almeida, and Mark Turner. 2024{\natexlab{a}}.
\newblock \emph{Copilots for Linguists: AI, Constructions, and Frames}.
\newblock Elements in Construction Grammar. Cambridge University Press, Cambridge.

\bibitem[{Torrent et~al.(2022)Torrent, Matos, Belcavello, Viridiano, Gamonal, Costa, and Marim}]{torrent-et-al-2022}
Tiago~Timponi Torrent, Ely Edison da~Silva Matos, Frederico Belcavello, Marcelo Viridiano, Maucha~Andrade Gamonal, Alexandre Diniz~da Costa, and Mateus~Coutinho Marim. 2022.
\newblock \href {https://doi.org/10.3389/fpsyg.2022.838441} {Representing {C}ontext in {F}rame{N}et: A {M}ultidimensional, {M}ultimodal {A}pproach}.
\newblock \emph{Frontiers in Psychology}, 13.

\bibitem[{Torrent et~al.(2024{\natexlab{b}})Torrent, Matos, Costa, Gamonal, Peron-Corr{\^e}a, and Paiva}]{torrent2024flexible}
Tiago~Timponi Torrent, Ely Edison da~Silva Matos, Alexandre Diniz~da Costa, Maucha~Andrade Gamonal, Simone Peron-Corr{\^e}a, and Vanessa Maria Ramos~Lopes Paiva. 2024{\natexlab{b}}.
\newblock \href {https://link.springer.com/article/10.1007/s10579-023-09714-8} {A flexible tool for a qualia-enriched {F}rame{N}et: the {F}rame{N}et {B}rasil {W}eb{T}ool}.
\newblock \emph{Language Resources and Evaluation}, pages 1--29.

\bibitem[{Trott et~al.(2020)Trott, Torrent, Chang, and Schneider}]{trott-etal-2020-construing}
Sean Trott, Tiago~Timponi Torrent, Nancy Chang, and Nathan Schneider. 2020.
\newblock \href {https://doi.org/10.18653/v1/2020.acl-main.462} {({R}e)construing meaning in {NLP}}.
\newblock In \emph{Proceedings of the 58th Annual Meeting of the Association for Computational Linguistics}, pages 5170--5184, Online. Association for Computational Linguistics.

\bibitem[{Viridiano et~al.(2024)Viridiano, Lorenzi, Timponi~Torrent, Matos, Pagano, Sathler~Sigiliano, Gamonal, de~Andrade~Abreu, Vicente~Dutra, Samagaio, Carvalho, Campos, Azalim, Mazzei, Fonseca~de Oliveira, Luz, Padua~Ruiz, Bellei, Pestana, Costa, Rabelo, Silva, Roza, Souza~Mota, Oliveira, and Pelegrino~de Freitas}]{viridiano-etal-2024-framed}
Marcelo Viridiano, Arthur Lorenzi, Tiago Timponi~Torrent, Ely~E. Matos, Adriana~S. Pagano, Nat{\'a}lia Sathler~Sigiliano, Maucha Gamonal, Helen de~Andrade~Abreu, L{\'i}via Vicente~Dutra, Mairon Samagaio, Mariane Carvalho, Franciany Campos, Gabrielly Azalim, Bruna Mazzei, Mateus Fonseca~de Oliveira, Ana~Carolina Luz, Livia Padua~Ruiz, J{\'u}lia Bellei, Amanda Pestana, Josiane Costa, Iasmin Rabelo, Anna~Beatriz Silva, Raquel Roza, Mariana Souza~Mota, Igor Oliveira, and M{\'a}rcio~Henrique Pelegrino~de Freitas. 2024.
\newblock \href {https://aclanthology.org/2024.lrec-main.656/} {Framed {M}ulti30{K}: A frame-based multimodal-multilingual dataset}.
\newblock In \emph{Proceedings of the 2024 Joint International Conference on Computational Linguistics, Language Resources and Evaluation (LREC-COLING 2024)}, pages 7438--7449, Torino, Italia. ELRA and ICCL.

\bibitem[{Xia et~al.(2021)Xia, Qin, Vashishtha, Chen, Chen, May, Harman, Rawlins, White, and Van~Durme}]{xia2021lome}
Patrick Xia, Guanghui Qin, Siddharth Vashishtha, Yunmo Chen, Tongfei Chen, Chandler May, Craig Harman, Kyle Rawlins, Aaron~Steven White, and Benjamin Van~Durme. 2021.
\newblock \href {https://doi.org/10.18653/v1/2021.eacl-demos.19} {{LOME}: Large ontology multilingual extraction}.
\newblock In \emph{Proceedings of the 16th Conference of the European Chapter of the Association for Computational Linguistics: System Demonstrations}, pages 149--159, Online. Association for Computational Linguistics.

\end{thebibliography}


\end{document}